\documentclass[10pt,twocolumn,letterpaper]{article}

\usepackage{cvpr}              

\usepackage{graphicx}
\usepackage{amsmath}
\usepackage{amssymb}
\usepackage{booktabs}
\usepackage{bm}
\usepackage{multirow} 
\usepackage{algorithm}
\usepackage{algorithmic}
\usepackage{subcaption}
\usepackage{enumitem}
\usepackage{siunitx}
\usepackage[nopar]{lipsum}
\usepackage[export]{adjustbox}
\usepackage[pagebackref,breaklinks,colorlinks]{hyperref}

\usepackage[capitalize]{cleveref}
\crefname{section}{Sec.}{Secs.}
\Crefname{section}{Section}{Sections}
\Crefname{table}{Table}{Tables}
\crefname{table}{Tab.}{Tabs.}

\def\cvprPaperID{1772} 
\def\confName{CVPR}
\def\confYear{2022}

\begin{document}

\title{Cross-Image Relational Knowledge Distillation for Semantic Segmentation}

\author{Chuanguang Yang$^{1,2}$ \qquad  Helong Zhou$^3$ \qquad Zhulin An$^1$\thanks{Corresponding author.}  \qquad Xue Jiang$^4$  \\ 
	Yongjun Xu$^1$   \qquad Qian Zhang$^3$ \\
	$^1$Institute of Computing Technology, Chinese Academy of Sciences, Beijing, China  \\ $^2$University of Chinese Academy of Sciences, Beijing, China \\  $^3$Horizon Robotics \qquad
	$^4$School of Computer Science, Wuhan University\\
	{\tt\small \{yangchuanguang, anzhulin, xyj\}@ict.ac.cn}
	\\ 
	{\tt\small
		\{helong.zhou, qian01.zhang\}@horizon.ai \quad jxt@whu.edu.cn
	}
}

\maketitle
\begin{abstract}
   Current Knowledge Distillation (KD) methods for semantic segmentation often guide the student to mimic the teacher's structured information generated from individual data samples. However, they ignore the global semantic relations among pixels across various images that are valuable for KD. This paper proposes a novel Cross-Image Relational KD (CIRKD), which focuses on transferring structured pixel-to-pixel and pixel-to-region relations among the whole images. The motivation is that a good teacher network could construct a well-structured feature space in terms of global pixel dependencies. CIRKD makes the student mimic better structured semantic relations from the teacher, thus improving the segmentation performance. Experimental results over Cityscapes, CamVid and Pascal VOC datasets demonstrate the effectiveness of our proposed approach against state-of-the-art distillation methods. The code is available at \url{https://github.com/winycg/CIRKD}.
\end{abstract}

\section{Introduction}
Semantic segmentation is a crucial and challenging task in computer vision. It aims to classify each pixel in the input image with an individual category label. The applications of segmentation often focus on autonomous driving, virtual reality and robots. Although popular state-of-the-art segmentation networks, such as DeepLab~\cite{chen2017deeplab,chen2018encoder}, PSPNet~\cite{zhao2017pyramid} and OCRNet~\cite{yuan2020object}, achieve remarkable performance, they often need high computational costs. This weakness makes them difficult to be deployed for real-world scenarios over resource-limited mobile devices. Therefore, a series of lightweight segmentation networks are proposed, such as ESPet~\cite{mehta2018espnet}, ICNet~\cite{zhao2018icnet} and BiSeNet~\cite{yu2018bisenet}. Moreover, model compression is also an alternative field to pursue compact networks, mainly divided into quantization~\cite{wu2016quantized}, pruning~\cite{yang2019multi,cai2022prior} and knowledge distillation (KD)~\cite{hinton2015distilling,yang2021hierarchical,shu2021channel}.

\begin{figure}[t]
	\centering 
	
	\begin{subfigure}[t]{0.22\textwidth}
		\centering
		\includegraphics[width=\textwidth]{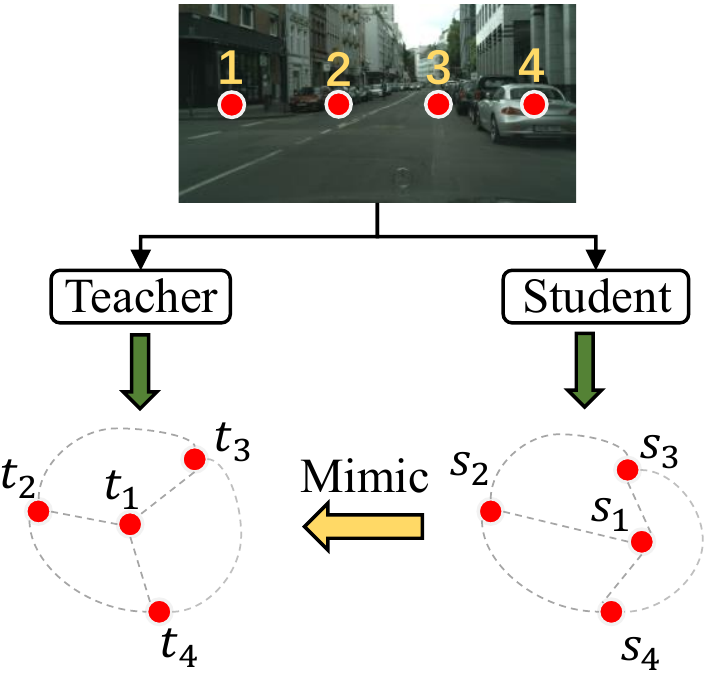}
		\caption{Intra-image relational KD.}
		\label{intra}
	\end{subfigure}
	\begin{subfigure}[t]{0.252\textwidth}
		\centering
		\includegraphics[width=\textwidth]{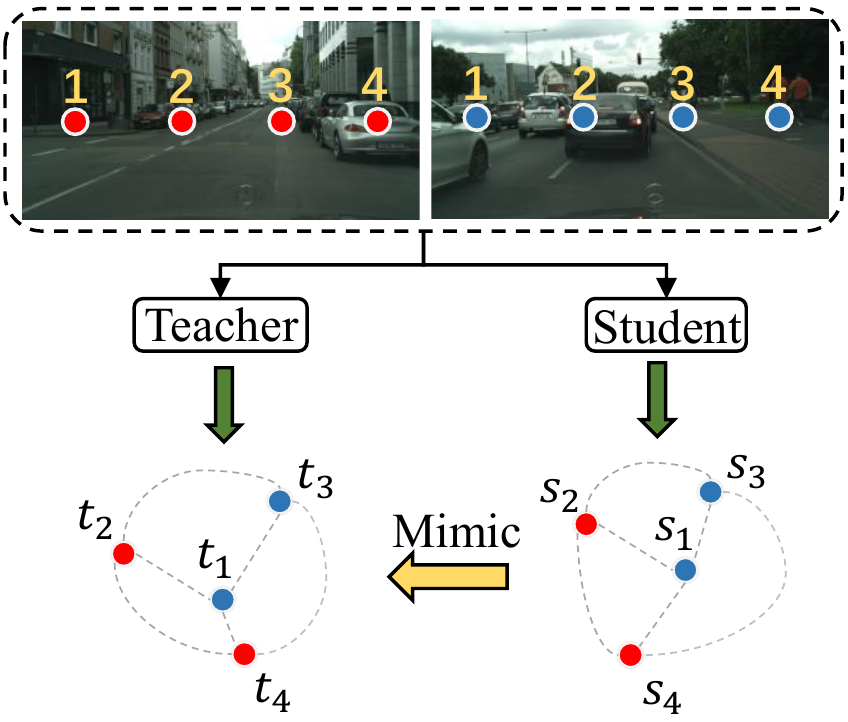}
		\caption{Cross-image relational KD.}
		\label{cross}
	\end{subfigure}
	\caption{Overview of intra-image (\emph{left}) and our proposed cross-image relational distillation (\emph{right}). The circles (\protect\includegraphics[scale=0.008]{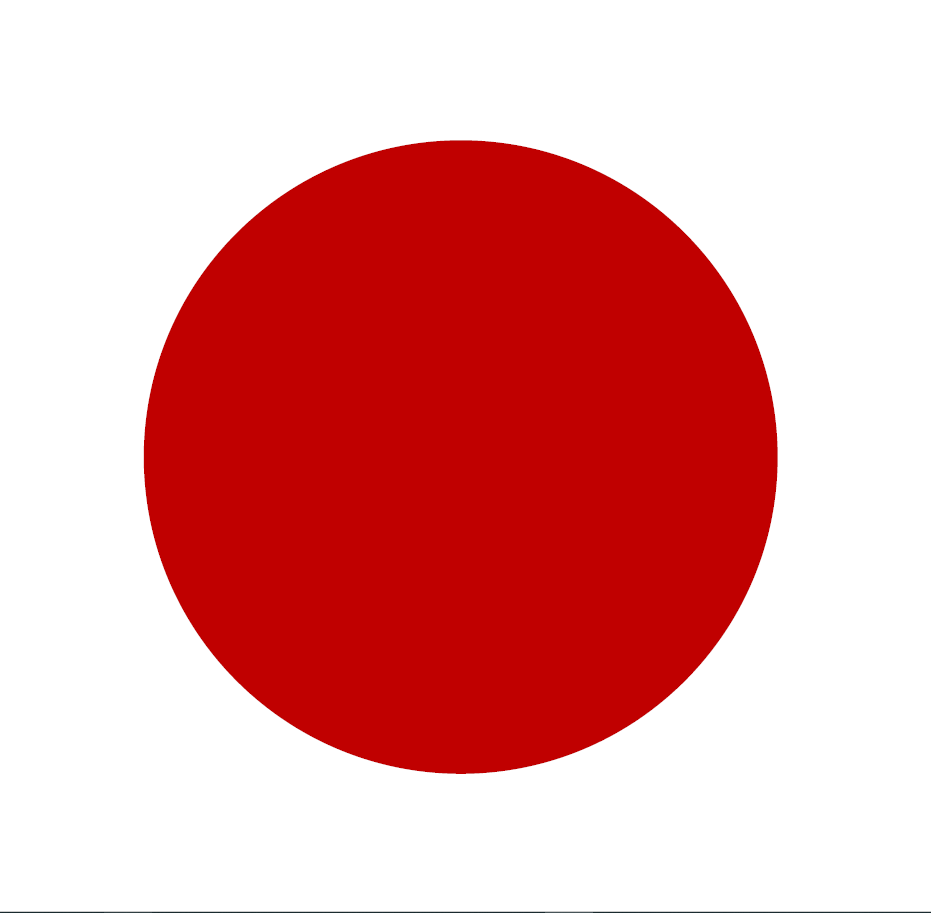} or \protect\includegraphics[scale=0.008]{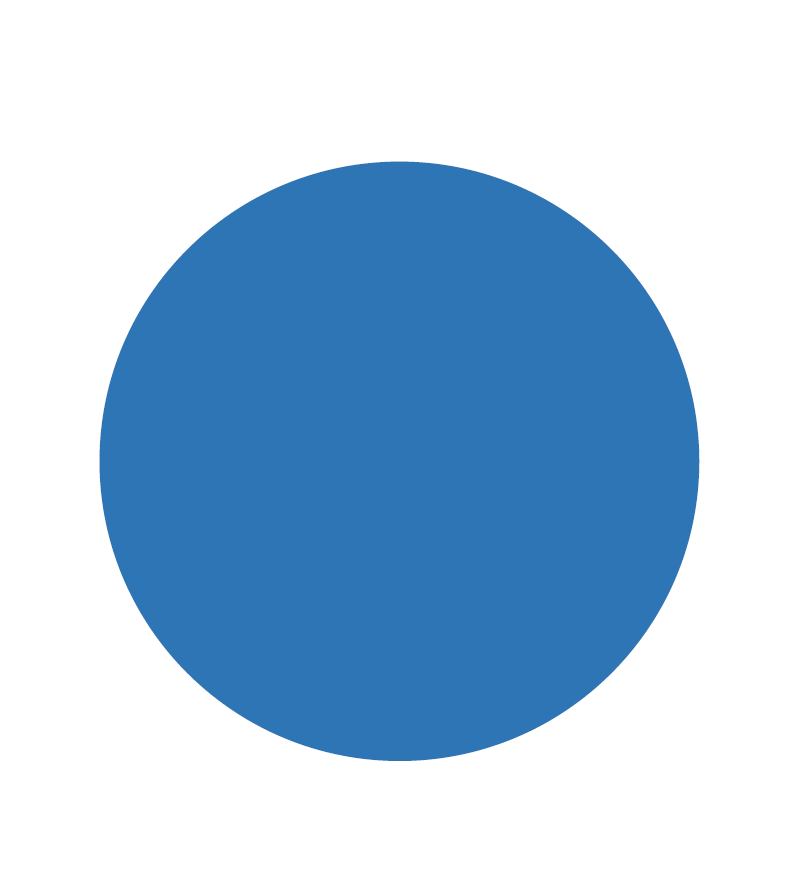}) with the same color denote pixel embeddings from the identical image. $t_{i}$ and $s_{i}$ represent the pixel embeddings of the $i$-th pixel location tagged in an image from the teacher and student, respectively. The dotted line (\protect\includegraphics[valign=c,scale=0.5]{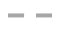}) shows the similarity relationship between two pixels. The circles and lines construct a relational graph.} 
	\label{introduction} 
	\vspace{-0.2cm}
\end{figure}

This paper investigates KD to improve the performance of a compact student network under the guidance of a high-capacity teacher network for semantic segmentation. A broad range of KD approaches~\cite{hinton2015distilling,li2017mimicking,zagoruyko2016paying,yang2021hierarchical}  have been well studied but mostly for image classification tasks. Unlike image-level recognition, the segmentation task aims at dense pixel predictions, which is more challenging. Previous researches~\cite{li2017mimicking,liu2020structured} have found that directly utilizing classification-based KD methods to deal with dense prediction tasks may not achieve desirable performance. This is because strictly aligning the coarse feature maps between the teacher and student networks may lead to negative constraints and ignore the structured context among pixels.

Recent works attempt to propose specialized KD methods~\cite{xie2018improving,liu2019structured,he2019knowledge,wang2020intra,liu2020efficient,shu2021channel} for semantic segmentation. Most focus on mining correlations or dependencies among spatial pixel locations because segmentation needs a structured output. Typical knowledge can be local pixel affinity~\cite{xie2018improving}, global pairwise relations~\cite{liu2019structured,he2019knowledge} and intra-class pixel variation~\cite{wang2020intra}. Such methods often perform better than the traditional point-wise alignment in capturing structured spatial knowledge. More recently, Shu \emph{et al.}~\cite{shu2021channel} revealed that each channel represents a category-specific mask and thus proposed Channel-Wise KD (CWD)~\cite{shu2021channel}. CWD achieves state-of-the-art distillation performance and demonstrates the importance of channel-level information for dense prediction tasks. However, previous segmentation KD methods often guide a student to mimic the teacher's structured information generated from \emph{individual data samples}. They ignore cross-image semantic relations among pixels for knowledge transfer, as shown in Fig.~\ref{introduction}.

Based on this motivation, we propose Cross-Image Relational Knowledge Distillation (CIRKD) for semantic segmentation. The core idea is to construct global pixel relations across the whole training images as meaningful knowledge. A good pre-trained teacher network could often generate a well-structured pixel embedding space and capture better pixel correlations than a student network. Based on this property, we transfer such pixel relations from teacher to student. Specifically, we propose pixel-to-pixel distillation and pixel-to-region distillation to fully exploit structured relations across various images. The former aims to transfer similarity distributions among pixel embeddings. The latter focuses on transferring pixel-to-region similarity distributions complementary to the former. The region embedding is generated by averagely pooling pixel embeddings from the same class and represents that class's feature center. The pixel-to-region relations indicate the relative similarities between pixels and class-wise prototypes.

A naive way for constructing cross-image relations is to derive embeddings from the current mini-batch. However, the batch size of the segmentation task is often small, limiting the network to capture broader pixel dependencies. Motivated by previous self-supervised learning~\cite{wu2018unsupervised,tian2020contrastive}, we introduce a pixel queue and a region queue in the memory bank to store abundant embeddings for modelling long-range pixel relations. The embeddings in queues are consistent during the distillation process, since they are generated from the pre-trained and frozen teacher network. We regard the teacher and student pixel embeddings from the current mini-batch as anchors. We randomly sample contrastive embeddings from the queues to model pixel-to-pixel as well as pixel-to-region similarity distributions. Then we align such soft relations via KL-divergence from the student to teacher.

 CIRKD guides the student network to learn the global property of relative pixel structures across training images from the teacher, further improving the segmentation performance. We evaluate our method over popular DeepLabV3~\cite{chen2018encoder} and PSPNet~\cite{zhao2017pyramid} architectures on three segmentation benchmark datasets: Cityscapes~\cite{cordts2016cityscapes}, CamVid~\cite{brostow2008segmentation}  and Pascal VOC~\cite{everingham2010pascal}. Experimental results indicate that CIRKD outperforms other state-of-the-art distillation approaches, demonstrating the value of transferring global pixel relationships in semantic segmentation.

The main contributions are summarized as follows:
\begin{itemize}[noitemsep,nolistsep,,topsep=0pt,parsep=0pt,partopsep=0pt]
	\item We propose cross-image relational KD to transfer global pixel relationships. We may be the first to build pixel dependencies across global images for segmentation KD.
	\item We propose pixel-to-pixel and pixel-to-region distillation with the memory bank mechanism to fully explore structured relations for transfer.
	\item Our CIRKD achieves the best distillation performance among state-of-the-art methods on the public segmentation datasets.
\end{itemize}

\section{Related Work}
\textbf{Semantic Segmentation.} Fully Convolutional Networks (FCN)~\cite{long2015fully} creates a seminal paradigm for end-to-end dense feature learning for semantic segmentation. Since contextual pixel dependencies are essential for segmentation performance~\cite{wang2004image}, capturing long-range relationships becomes a critical topic. DeepLab~\cite{chen2017deeplab} applies atrous convolution to enlarge the receptive field for learning broader context. DeepLabV3~\cite{chen2017rethinking} assembles convolution blocks with various atrous rates in parallel to capture multi-scale contexts. PSPNet~\cite{zhao2017pyramid} proposes a pyramid pooling module to exploit different-region-based context aggregation. RefineNet~\cite{lin2017refinenet} preserves high-resolution predictions by long-range residual connections for the down-sampling process. More recently, SegFormer~\cite{xie2021segformer} utilizes a structured Transformer encoder to model global context information. However, such high-performance segmentation networks with expensive computational costs are difficult to be deployed over resource-limited mobile devices. 

Efficient segmentation networks attract wide attention due to the need for real-time inference. Most works attempt to design lightweight networks with cheap operations. ENet~\cite{paszke2016enet} is equipped with early downsampling, small decoder size and filter factorization.  ESPNet~\cite{mehta2018espnet} factorizes the standard convolution into the spatial pyramid of dilated convolution. ICNet~\cite{zhao2018icnet} builds a cascade structure to balance the efficiency between low-resolution and high-resolution features. BiSeNet~\cite{yu2018bisenet} combines a spatial path and a context path to process features efficiently. Beyond designing a segmentation framework, lightweight backbone networks~\cite{sandler2018mobilenetv2,zhang2018shufflenet,yang2020gated}, \emph{e.g.} MobileNet~\cite{sandler2018mobilenetv2} and ShuffleNet~\cite{zhang2018shufflenet}, can also implement acceleration.

\textbf{Knowledge Distillation.} The core idea of KD is to transfer meaningful knowledge from a cumbersome teacher into a smaller and faster student. Most current KD methods deal with image classification networks, mainly divided into probability-based, feature-based and relation-based approaches. Probability-based KD~\cite{hinton2015distilling,zhou2021rethinking} transfers class probabilities produced from the teacher as soft labels to supervise the student. Feature-based KD focuses on intermediate feature maps~\cite{romero2014fitnets} or their refined information~\cite{zagoruyko2016paying,heo2019comprehensive} as knowledge. Relation-based KD~\cite{peng2019correlation,park2019relational,tung2019similarity,fang2021seed,yang2021multi,yang2022mutual} aligns correlations or dependencies among multiple instances between the student and teacher networks. Our CIRKD is related to SEED~\cite{fang2021seed} that both of them are contrastive distillation manners with a shared memory bank. However, these image-level KD methods are often unsuitable for pixel-wise semantic segmentation~\cite{li2017mimicking,liu2020structured}.

Recent KD methods for semantic segmentation often encode contextual pixel affinity as knowledge. Xie \emph{et al.}~\cite{xie2018improving} align local similarity maps constructed from 8 neighbourhood pixels between the student and teacher networks. He \emph{et al.}~\cite{he2019knowledge} transfer non-local pairwise affinity maps with an autoencoder to minimize the discrepancy of features. Liu \emph{et al.}~\cite{liu2019structured,liu2020structured} perform a pairwise similarity distillation among pixels and an adversarial distillation of score maps. Wang \emph{et al.}~\cite{wang2020intra} distill intra-class feature variation to learn more robust relations with class-wise prototypes. Beyond spatial distillation, Shu \emph{et al.}~\cite{shu2021channel} propose channel-wise distillation to guide the student to mimic the teacher's semantic masks along the channel dimension. Though achieving desirable performance, these approaches only consider pixel dependencies within an individual image, ignoring global pixel relations across various images.

\section{Methodology}
\subsection{Preliminary}
\textbf{Notations of Semantic Segmentation Framework.} Unlike traditional image classification, semantic segmentation is a pixel-wise dense classification task. A segmentation network needs to classify each pixel in the image to an individual category label from $C$ classes. The network can be decomposed of a feature extractor and a classifier. The former generates a dense feature map $\mathbf{F}\in \mathbb{R}^{H\times W\times d}$, where $H$, $W$ and $d$ denote the height, width and number of channels, respectively. We can derive $H\times W$ pixel embeddings along the spatial dimension. The latter further transforms the $\mathbf{F}$ into a categorical logit map $\mathbf{Z}\in \mathbb{R}^{H\times W\times C}$. The conventional segmentation task loss is to train each pixel with its ground-truth label using cross-entropy:
\begin{equation}
	L_{task}=\frac{1}{H\times W}\sum_{h=1}^{H}\sum_{w=1}^{W} CE(\sigma(\mathbf{Z}_{h,w}),y_{h,w}).
	\label{task}
\end{equation}
Here, $CE$ denotes the cross-entropy loss, $\sigma$ denotes the softmax function and $y_{h,w}$ denotes the ground-truth label of the $(h,w)$-th pixel.

\textbf{Pixel-wise Class Probability Distillation.} Motivated by Hinton's KD~\cite{hinton2015distilling}, a direct method is to align the class probability distribution of each pixel from the student to the teacher. The formulation is expressed as:
\begin{equation}
	L_{kd}=\frac{1}{H\times W}\sum_{h=1}^{H}\sum_{w=1}^{W} KL(\sigma(\frac{\mathbf{Z}^{s}_{h,w}}{T})||\sigma(\frac{\mathbf{Z}^{t}_{h,w}}{T})).
	\label{kd}
\end{equation}
Here, $\sigma(\mathbf{Z}^{s}_{h,w}/T)$ and $\sigma(\mathbf{Z}^{t}_{h,w}/T)$ represent the soft class probabilities of the $(h,w)$-th pixel produced from the student and teacher, respectively.  $KL$ denotes the Kullback-Leibler divergence, and $T$ is a temperature. Following previous works~\cite{liu2019structured,wang2020intra}, $T=1$ is good enough.

\subsection{Cross-Image Relational Knowledge Distillaton}
\textbf{Motivation.} Although the training objectives of $L_{task}$ and $L_{kd}$ are widely used in semantic segmentation, they only deal with pixel-wise predictions independently but neglect semantic relations between pixels. Some segmentation KD methods~\cite{liu2019structured,he2019knowledge,wang2020intra} attempt to capture spatial  relational knowledge by modelling pixel affinity. Nevertheless, these KD methods
only construct the relationships among pixels \emph{within a single image}, regardless of the semantic dependencies among pixels across global images.  This paper demonstrates that cross-image relational knowledge is also valuable for conducting teacher-student-based KD.

Our CIRKD makes use of pixel embeddings beyond a single image. Inspired by the recent memory-based contrastive learning~\cite{wu2018unsupervised,tian2020contrastive,wang2020cross,chen2020simple}, we may retrieve pixel embeddings of other images from \emph{the current mini-batch} or \emph{an online memory bank}. This paper considers both of two manners to model relationships among pixels, the details of which are shown as follows. 
\begin{figure*}[tbp]  
	\centering 
	\includegraphics[width=1.\linewidth]{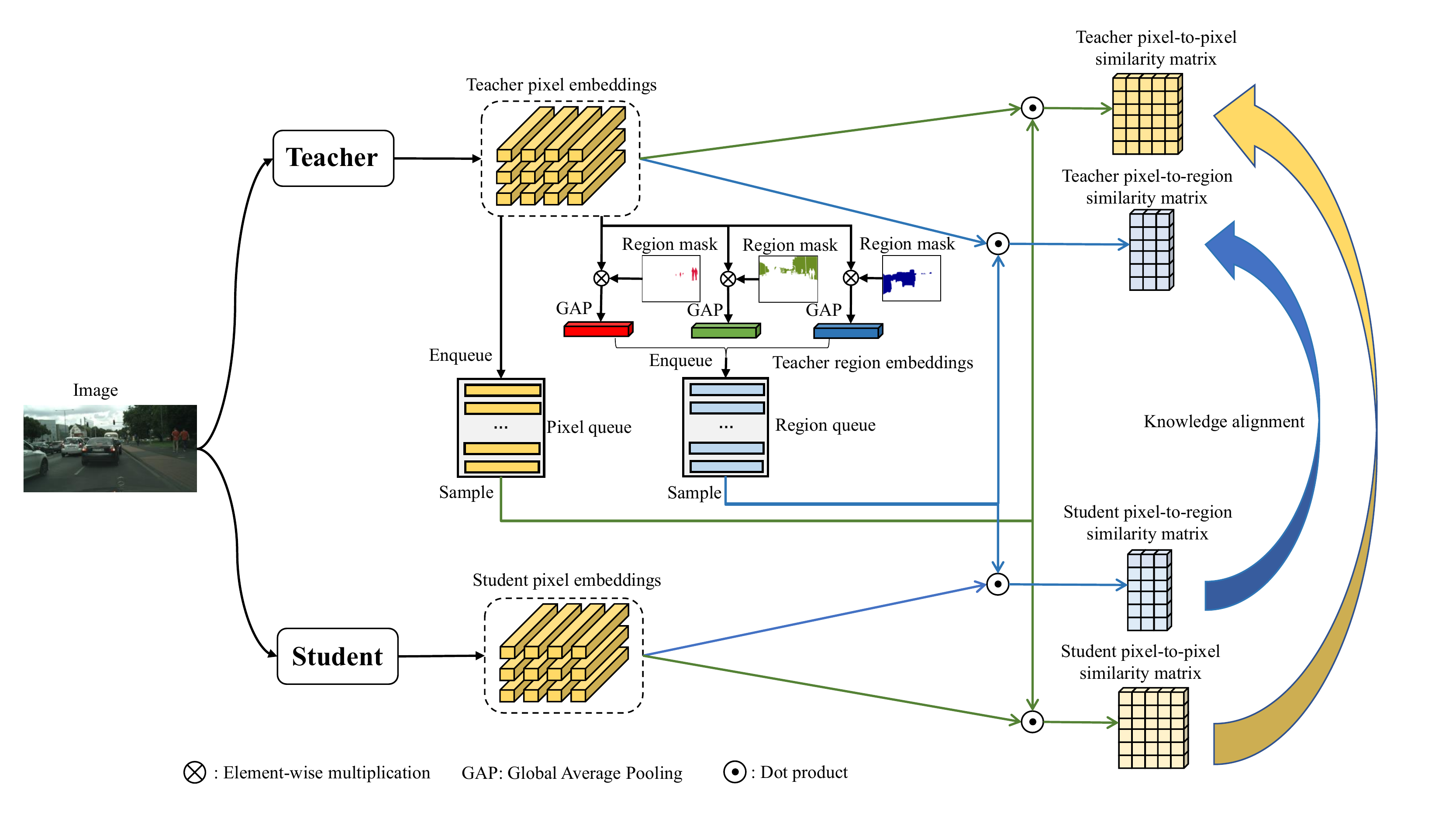}
	\caption{Overview of our proposed memory-based pixel-to-pixel distillation and pixel-to-region distillation.} 
	\label{mem_kd}
	\vspace{-0.2cm}
\end{figure*}
\subsubsection{Mini-batch-based Pixel-to-Pixel Distillation}

\label{minibatch}

 Given a mini-batch $\{\bm{x}_{n}\}_{n=1}^{N}$, the segmentation network extracts $N$ structured feature maps $\{\mathbf{F}_{n}\in \mathbb{R}^{H\times W\times d}\}_{n=1}^{N}$ from $N$ input images. We preprocess each pixel embedding in $\mathbf{F}_{n}$ by $l_{2}$-normalization. For easy notation, we reshape the spatial dimension of $\{\mathbf{F}_{n}\in \mathbb{R}^{H\times W\times d}\}_{n=1}^{N}$ to $\{\mathbf{F}_{n}\in \mathbb{R}^{A\times d}\}_{n=1}^{N}$, where $A=H\times W$. For the $i$-th image $\bm{x}_{i}$ and the $j$-th image $\bm{x}_{j}$, $i,j\in \{1,2,\cdots,N\}$, we can calculate the cross-image pair-wise similarity matrix $\mathbf{S}_{ij}=\mathbf{F}_{i}\mathbf{F}_{j}^\top\in \mathbb{R}^{A\times A}$.
The relational matrix $\mathbf{S}_{ij}$ captures the cross-image pair-wise correlations among pixels. 

We guide the pair-wise similarity matrix of $\mathbf{S}^{s}_{ij}$ produced from the student to align that of $\mathbf{S}^{t}_{ij}$ produced from the teacher. The distillation process is formulated as:
\begin{equation}
L_{p2p}(\mathbf{S}^{s}_{ij},\mathbf{S}^{t}_{ij})=\frac{1}{A}\sum_{a=1}^{A}KL(\sigma(\frac{\mathbf{S}^{s}_{ij|a,:}}{\tau})||\sigma(\frac{\mathbf{S}^{t}_{ij|a,:}}{\tau})).
\end{equation}
Here, $\mathbf{S}_{ij|a,:}$ denotes the $a$-th row vector of $\mathbf{S}_{ij}$. We normalize each row similarity distribution of $\mathbf{S}_{ij}$ to a probability distribution with a temperature $\tau$ by softmax function $\sigma$. The magnitude gaps would be removed between the student and teacher networks due to the softmax normalization. $KL$ is used to align each row-wise probability distribution. We perform pixel-to-pixel distillation every two of $N$ images:
\begin{equation}
	L_{batch\_p2p}=\frac{1}{N^{2}}\sum_{i=1}^{N}\sum_{j=1}^{N}L_{p2p}(\mathbf{S}^{s}_{ij},\mathbf{S}^{t}_{ij}).
\end{equation}
We show the illustration of mini-batch-based pixel-to-pixel distillation in the supplement.

\subsubsection{Memory-based Pixel-to-Pixel Distillation}
Although mini-batch-based distillation could capture cross-image relations to some extent, it is difficult to model dependencies among pixels from global images, since the batch size per GPU of segmentation task is often small, \emph{e.g.} 1 or 2. To address this problem, we introduce an online pixel queue that can store massive pixel embeddings in the memory bank generated from the past mini-batches. It allows us to retrieve abundant embeddings efficiently. The usage of memory bank dates back to self-supervised learning~\cite{wu2018unsupervised,tian2020contrastive}. This is because a large number of negative samples are pivotal for unsupervised contrastive learning, and the mini-batch size limits available contrastive samples. 

In the context of the dense segmentation task, each image would contain a vast number of pixel samples, and most pixels in the same object region are often homogeneous. Therefore storing all pixel embeddings may learn redundant relational knowledge and slow down the distillation process. Moreover, saving several last batches to the queue may also damage the diversity of pixel embeddings. Thus we maintain a class-aware pixel queue $\mathcal{Q}_{p}\in \mathbb{R}^{C\times N_{p}\times d}$, where $N_{p}$ is the number of pixel embeddings per class and $d$ is the embedding size. For each image in the mini-batch, we only randomly sample a small number, \emph{i.e.} $V$ ($V\ll N_{p}$), of pixel embeddings from the same class and push them into the pixel queue $\mathcal{Q}_{p}$. The queue is progressively
updated under the “first-in-first-out” strategy as distillation proceeds. 

Inspired by~\cite{fang2021seed}, we adopt a shared pixel queue between the student and teacher networks and store pixel embeddings generated from the teacher during the distillation process. Given an input image $\bm{x}_{n}$, the generated pixel embeddings of the student and teacher networks are $\mathbf{F}_{n}^{s}\in \mathbb{R}^{A\times d}$ and $\mathbf{F}_{n}^{t}\in \mathbb{R}^{A\times d}$, respectively. Each pixel embedding of $\mathbf{F}_{n}^{s}$ and $\mathbf{F}_{n}^{t}$ is preprocessed by $l_{2}$-normalization. We regard $\mathbf{F}_{n}^{s}$ and $\mathbf{F}_{n}^{t}$ as anchors and sample $K_{p}$ contrastive embeddings $\{\bm{v}_{k}\in \mathbb{R}^{d}\}_{k=1}^{K_{p}}$ randomly from the pixel queue $\mathcal{Q}_{p}$. Here, we adopt a class-balanced sampling since the numbers of pixels from various classes often conform to a long-tailed distribution. For easy notation, $\mathbf{V}_{p}=[\bm{v}_{1},\bm{v}_{2},\cdots,\bm{v}_{K_{p}}]\in \mathbb{R}^{K_{p}\times d}$ is a concatenation of $\{\bm{v}_{k}\}_{k=1}^{K_{p}}$ along the row dimension.  Then we model the pixel similarity matrix between the anchors and contrastive embeddings for the student and teacher as $\mathbf{P}^{s}$ and $\mathbf{P}^{t}$:
\begin{equation}
	\mathbf{P}^{s}=\mathbf{F}_{n}^{s}\mathbf{V}_{p}^{\top}\in \mathbb{R}^{A\times K_{p}},\  \mathbf{P}^{t}=\mathbf{F}_{n}^{t}\mathbf{V}_{p}^{\top}\in \mathbb{R}^{A\times K_{p}}.
\end{equation}

The teacher network often shows a better pixel similarity matrix than the student. We force the student's $\mathbf{P}^{s}$ to mimic the teacher's $\mathbf{P}^{t}$ for penalizing the difference. Similar to Section~\ref{minibatch}, we apply softmax normalization on each row distribution of $\mathbf{P}^{s}$ and $\mathbf{P}^{t}$ and perform pixel-to-pixel distillation via KL-divergence loss. It is formulated as follows:
\begin{equation}
L_{memory\_p2p}=\frac{1}{A}\sum_{a=1}^{A}KL(\sigma(\frac{\mathbf{P}^{s}_{a,:}}{\tau})||\sigma(\frac{\mathbf{P}^{t}_{a,:}}{\tau})).
\label{memory_p2p}
\end{equation}

After each iteration, we push $V$ teacher pixel embeddings per class into the pixel queue $\mathcal{Q}_{p}$. Because the teacher is pre-trained and frozen, it can provide consistent feature embeddings during the distillation process. Therefore, we can naturally avoid the inconsistent problem between the anchor and dequeued features appeared in previous contrastive learning~\cite{he2020momentum,hu2021region,wang2021exploring}.

\subsubsection{Memory-based Pixel-to-Region Distillation}
Discrete pixel embeddings may not fully capture image content. Thus we introduce an online region queue that can store massive more representative region embeddings in the memory bank.  Beyond pixel-to-pixel distillation, we further construct pixel-to-region distillation to model the relations between pixels and class-wise region embeddings across global images. Each region embedding represents the feature center of one semantic class in an image. We formulate the region embedding of class $c$ by averagely pooling all the pixel embeddings belonging to class $c$ in a single image. 

We maintain a region queue $\mathcal{Q}_{r}\in \mathbb{R}^{C\times N_{r}\times d}$ during the distillation process, where $N_{r}$ is the number of region embeddings per class and $d$ is the embedding size. For each iteration, we sample $K_{r}$ contrastive region embeddings $\{\bm{r}_{k}\in \mathbb{R}^{d}\}_{k=1}^{K_{r}}$ from $\mathcal{Q}_{r}$ in a class-balanced manner. For easy notation, $\mathbf{V}_{r}=[\bm{r}_{1},\bm{r}_{2},\cdots,\bm{r}_{K_{r}}]\in \mathbb{R}^{K_{r}\times d}$ is a concatenation of $\{\bm{r}_{k}\}_{k=1}^{K_{r}}$ along the row dimension. Given an input image $\bm{x}_{n}$, we model the pixel-to-region similarity matrix from $\mathbf{F}_{n}^{s}\in \mathbb{R}^{A\times d}$ and $\mathbf{F}_{n}^{t}\in \mathbb{R}^{A\times d}$ to region embeddings $\mathbf{V}_{r}$ as $\mathbf{R}^{s}$ and $\mathbf{R}^{t}$:
\begin{equation}
\mathbf{R}^{s}=\mathbf{F}_{n}^{s}\mathbf{V}_{r}^{\top}\in \mathbb{R}^{A\times K_{r}},\  \mathbf{R}^{t}=\mathbf{F}_{n}^{t}\mathbf{V}_{r}^{\top}\in \mathbb{R}^{A\times K_{r}}.
\end{equation}

Similar to the Equ.~(\ref{memory_p2p}), we distill normalized pixel-to-region similarity matrix between the student and teacher networks via KL-divergence loss:
\begin{equation}
	L_{memory\_p2r}=\frac{1}{A}\sum_{a=1}^{A}KL(\sigma(\frac{\mathbf{R}^{s}_{a,:}}{\tau})||\sigma(\frac{\mathbf{R}^{t}_{a,:}}{\tau})).
	\label{memory_p2r}
\end{equation}
For each mini-batch, we push all teacher region embeddings into the region queue $\mathcal{Q}_{r}$. The overview of our proposed memory-based distillation is shown in Fig.~\ref{mem_kd}.

\subsection{Overall Framework}
We summarize our mini-batch-based pixel-to-pixel, memory-based pixel-to-pixel and pixel-to-region distillation together to train the student network. We also employ the conventional pixel-wise cross-entropy task loss $L_{task}$ (Equ.~(\ref{task})) and class probability KD loss $L_{kd}$ (Equ.~(\ref{kd})) as the basic losses. The overall loss is formulated as:
\begin{align}
	L_{CIRKD}=&L_{task}+L_{kd}+\alpha L_{batch\_p2p} \notag \\
	&+\beta L_{memory\_p2p}+\gamma L_{memory\_p2r}.
	\label{overall_loss}
\end{align}
Here, $\alpha$, $\beta$ and $\gamma$ are weights coefficients. We set $\alpha=1$, $\beta=0.1$ and $\gamma=0.1$. Empirically, we find our CIRKD are not sensitive to coefficients when $\alpha,\beta,\gamma \in [0.1,1]$. When the student and teacher networks mismatch the embedding size, we attach a projection head to the student network.  It can map the student's pixel embeddings to match the teacher's dimension. The projection head is composed of two $1\times 1$ convolutional layers with ReLU and batch normalization. It would be discarded at the inference phase without introducing extra costs. In Algorithm \ref{cirkd_alg}, we use pseudo-code to illustrate the overall training pipeline of CIRKD.

\begin{algorithm}[tb]
	\caption{Cross-Image Relational KD (CIRKD)}
	\begin{algorithmic}
		\label{cirkd_alg}
		\STATE
		Initialize the pixel queue $\mathcal{Q}_{p}$ and the region queue $\mathcal{Q}_{r}$ with random unit vectors. 
		\WHILE{the student network has not converged}
		\STATE 	Sample a mini-batch.
		\STATE Generate the student and teacher pixel embeddings.
		\STATE  Compute the mini-batch-based pixel-to-pixel distillation loss $L_{batch\_p2p}$.
		\STATE 	Sample contrastive pixel and region embeddings from the pixel queue $\mathcal{Q}_{p}$ and the region queue $\mathcal{Q}_{r}$.
		\STATE Compute the memory-based pixel-to-pixel loss $L_{memory\_p2p}$ and pixel-to-region loss $L_{memory\_p2r}$.
		\STATE Update the student w.r.t the overall loss $L_{CIRKD}$.
		\STATE Enqueue the current teacher pixel and region embeddings to $\mathcal{Q}_{p}$ and $\mathcal{Q}_{r}$.
		\STATE Dequeue the earliest pixel and region embeddings from $\mathcal{Q}_{p}$ and $\mathcal{Q}_{r}$.
		\ENDWHILE
	\end{algorithmic}

\end{algorithm}

\section{Experiments}
\subsection{Experimental Setup}
\textbf{Dataset.} We employ three popular semantic segmentation datasets to conduct our experiments. (1) \textbf{Cityscapes}~\cite{cordts2016cityscapes} is an urban scene parsing dataset that contains 5000 finely annotated images, where 2975/500/1525 images are used for \texttt{train/val/test}. The segmentation performance is reported on 19 classes. (2) \textbf{CamVid}~\cite{brostow2008segmentation} is an automotive dataset that contains 367/101/233 images for \texttt{train/val/test} with 11 semantic classes. (3) \textbf{Pascal VOC}~\cite{everingham2010pascal} is a visual object segmentation dataset that includes 20 foreground object categories and one background class. We adopt the augmented data with extra annotations provided by~\cite{hariharan2011semantic}. The resulting dataset contains 10582/1449/1456 images for \texttt{train/val/test}.

\textbf{Evaluation metrics.} Following the standard setting, we employ mean
Intersection-over-Union (mIoU) to measure the segmentation performance.
\begin{figure*}[tbp]  
	\centering 
	\includegraphics[width=1\linewidth]{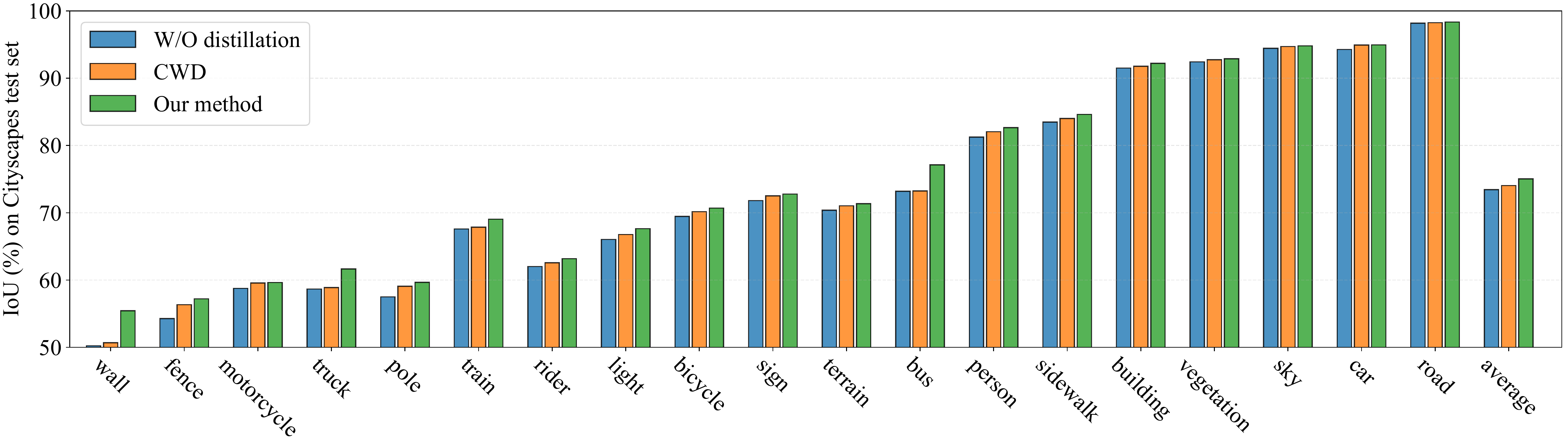}
	\caption{Illustration of individual class IoU scores over the student network DeepLabV3-ResNet18 with baseline (w/o distillation), state-of-the-art CWD and our proposed CIRKD on Cityscapes test set. Our CIRKD can consistently improve individual class IoU scores compared to the baseline and CWD, especially for those challenging classes with low IoU scores. } 
	\label{cityscape_iou}
	\vspace{-0.2cm}
\end{figure*}

\textbf{Network architectures.} For all experiments, we use the segmentation framework DeepLabV3~\cite{chen2018encoder} with ResNet-101 (Res101) backbone~\cite{he2016deep} as the powerful teacher network. For student networks, we use various segmentation architectures to verify the effectiveness of distillation methods. Specifically, DeepLabV3 and PSPNet~\cite{zhao2017pyramid} with different backbones of ResNet-18 (Res18) and MobileNetV2 (MBV2)~\cite{sandler2018mobilenetv2} are adopted.

\textbf{Training details.} Following the standard data augmentation, we employ random flipping and scaling in the range of $[0.5,2]$. All experiments are optimized by SGD with a momentum of 0.9, a batch size of 16 and an initial learning rate of 0.02. The number of the total training iterations is 40K. The learning rate is decayed by $(1-\frac{iter}{total\_iter})^{0.9}$ following the polynomial annealing policy~\cite{chen2017rethinking}. For crop size during the training phase, we use $512\times 1024$, $360\times 360$ and $512\times 512$ for Cityscapes, CamVid and Pascal VOC, respectively. 

\textbf{Evaluation details.} We evaluate the segmentation performance under a single scale setting over the original image size following the general protocol~\cite{shu2021channel}. 

\textbf{Compared distillation methods.} We compare our proposed CIRKD with state-of-the-art segmentation distillation methods: SKD~\cite{liu2019structured}, IFVD~\cite{wang2020intra} and CWD~\cite{shu2021channel}. We re-run all methods using author-provided code. All methods use the same pre-trained teacher DeepLabV3-ResNet101.

\textbf{Hyper-parameters setup.} The hyper-parameters are mainly from the pixel and region queues. For the pixel queue, we set $N_{p}=20K$ for each class and enqueue $V=16$ pixels per class for each image.  For the region queue, we set $N_{r}=2K$ for each class. For each mini-batch, we sample $K_{p}=4096$ pixel embeddings from the pixel queue and $K_{r}=1024$ region embeddings from the region queue to compute similarity matrices.

\begin{table}[tbp]
	\centering
	\resizebox{1.\linewidth}{!}{
		\begin{tabular}{l|c|c|ccc}  
			\hline
			\multirow{2}{*}{Method}&\multirow{2}{*}{Params (M)}&\multirow{2}{*}{FLOPs (G)}& \multicolumn{2}{c}{mIoU (\%)} \\ 
			&&&Val&Test \\
			\hline
			T: DeepLabV3-Res101 &61.1M &2371.7G &78.07 &77.46 \\
			\hline
			S: DeepLabV3-Res18 &\multirow{5}{*}{13.6M} &\multirow{5}{*}{572.0G} &74.21 & 73.45\\
			+SKD~\cite{liu2019structured} & & &75.42 & 74.06\\
			+IFVD~\cite{wang2020intra} & & & 75.59& 74.26\\
			+CWD~\cite{shu2021channel} & & & 75.55& 74.07\\
			+CIRKD (ours) & & &\textbf{76.38} & \textbf{75.05}\\
			\hline
			S: DeepLabV3-Res18* &\multirow{5}{*}{13.6M} &\multirow{5}{*}{572.0G} & 65.17 & 65.47\\
			+SKD~\cite{liu2019structured} & & & 67.08&66.71 \\
			+IFVD~\cite{wang2020intra} & & & 65.96& 65.78\\
			+CWD~\cite{shu2021channel} & & & 67.74& 67.35\\
			+CIRKD (ours) & & & \textbf{68.18}& \textbf{68.22} \\
			\hline
			S: DeepLabV3-MBV2 &\multirow{5}{*}{3.2M} &\multirow{5}{*}{128.9G} & 73.12&72.36 \\
			+SKD~\cite{liu2019structured} & & & 73.82&73.02 \\
			+IFVD~\cite{wang2020intra} & & & 73.50& 72.58\\
			+CWD~\cite{shu2021channel} & & &74.66 & 73.25\\
			+CIRKD (ours) & & & \textbf{75.42}&\textbf{74.03} \\
			\hline
			S: PSPNet-Res18 &\multirow{5}{*}{12.9M} &\multirow{5}{*}{507.4G} &72.55 &72.29 \\
			+SKD~\cite{liu2019structured} & & &73.29 &72.95 \\
			+IFVD~\cite{wang2020intra} & & & 73.71&72.83 \\
			+CWD~\cite{shu2021channel} & & &74.36 & 73.57\\
			+CIRKD (ours) & & &\textbf{74.73} &\textbf{74.05} \\
			\hline   
	\end{tabular}}
	\vspace{-0.1cm}
	\caption{Performance comparison with state-of-the-art distillation methods over various student segmentation networks on Cityscapes. * denotes that we do not initialize the backbone with ImageNet~\cite{deng2009imagenet} pre-trained weights. FLOPs is measured based on the fixed size of $1024\times 2048$. The bold number denotes the best result in each block. We tag the teacher as T and the student as S.}
	\label{citys} 
	\vspace{-0.5cm}
\end{table}

\begin{figure}[tbp]  
	\centering 
	\includegraphics[width=1.\linewidth]{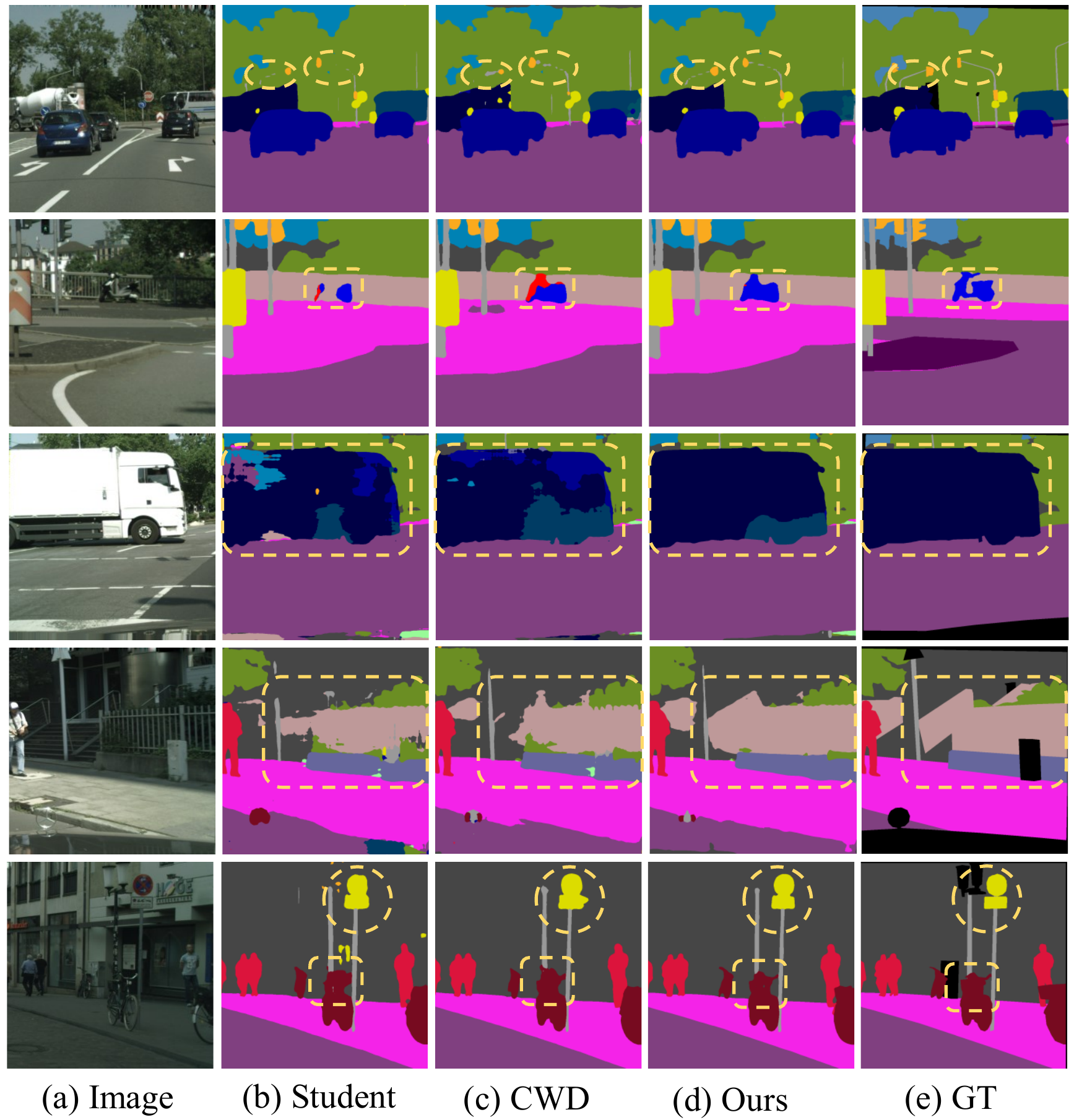}
	\caption{Qualitative segmentation results on the validation set of Cityscapes using
		the DeepLabV3-ResNet18 network: (a) raw images, (b) the
		original student network without KD,
		(c) channel-wise distillation, (d) our method  and (e) ground truth.} 
	\label{Qualitative}
	\vspace{-0.6cm}
\end{figure}

\subsection{Experimental Results}
\subsubsection{Results on Cityscapes} 
In Table~\ref{citys}, we compare our proposed CIRKD against state-of-the-art distillation methods on Cityscapes in terms of the validation and test mIoU performance. We can observe that all structured KD methods improve student networks under the teacher's supervision. CIRKD achieves the best segmentation performance across various student networks with similar or different architecture styles. It reveals that CIRKD does not rely on architecture-specific cues. Moreover, our method outperforms the best completing CWD with an average 0.60\% validation mIoU gain and 0.78\% test mIoU gain across four student networks. The results demonstrate that distilling cross-image relations guides the student to achieve better segmentation performance than intra-image pixel affinity~\cite{liu2019structured,wang2020intra}.
\begin{figure}[t]
	\centering 
	
	\begin{subfigure}[t]{0.235\textwidth}
		\centering
		\includegraphics[width=\textwidth]{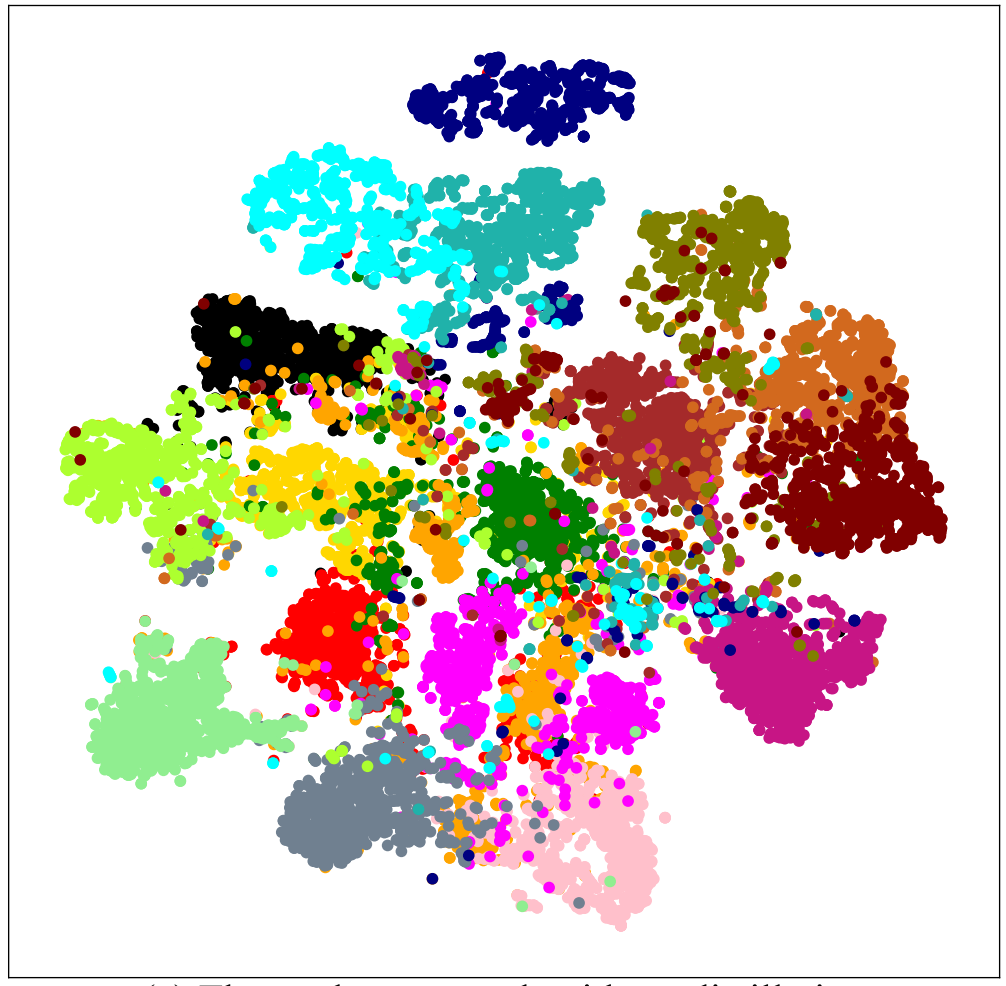}
		\caption{State-of-the-art CWD~\cite{shu2021channel}.}
		\label{SMG}
	\end{subfigure}
	\begin{subfigure}[t]{0.235\textwidth}
		\centering
		\includegraphics[width=\textwidth]{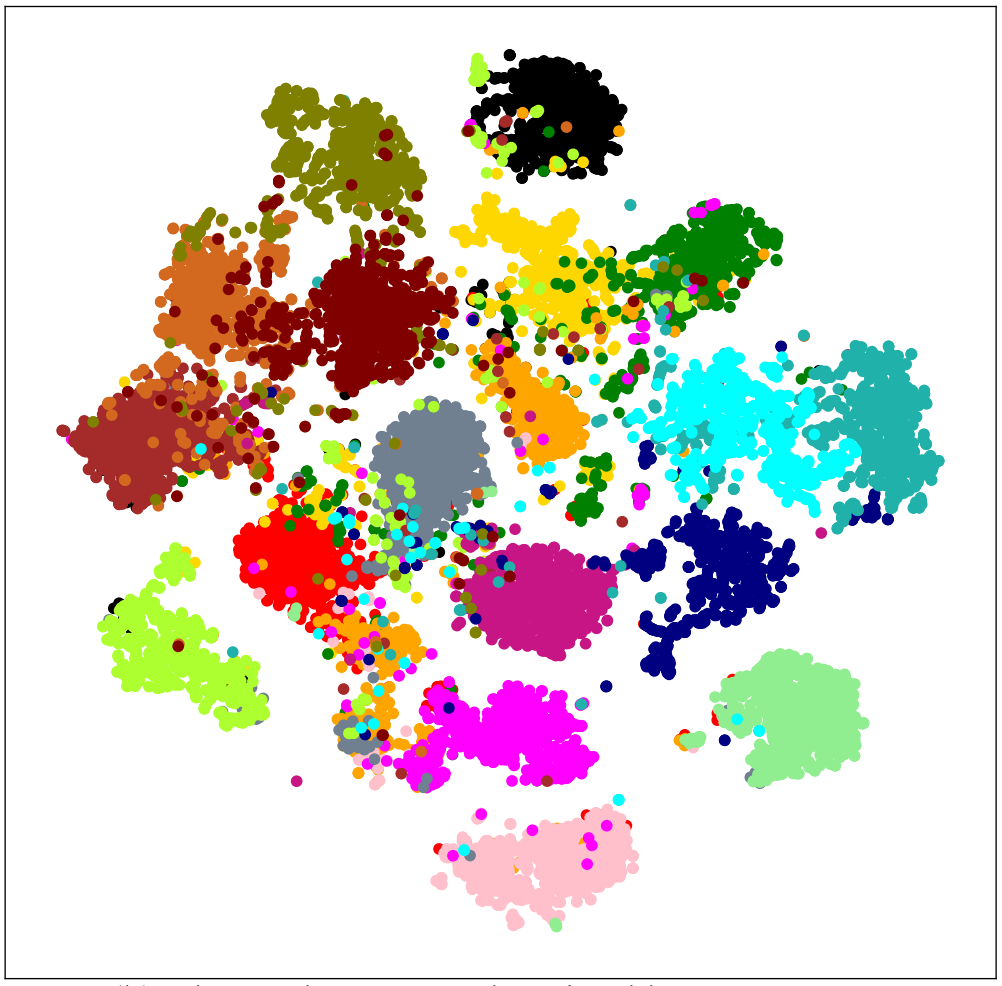}
		\caption{Our CIRKD.}
		\label{Update}
	\end{subfigure}
\vspace{-0.1cm}
	\caption{T-SNE visualization of learned features embeddings on the validation set of Cityscapes over the DeepLabV3-ResNet18 network trained with CWD (\emph{left}) and our proposed CIRKD (\emph{right}).} 
	\label{tsne} 
	
\end{figure} 

As illustrated in Fig.~\ref{cityscape_iou}, we also show the performance of individual class IoU scores over the student network. We can observe that our CIRKD achieves better class IoU scores than baseline (w/o distillation) and CWD consistently, especially for those categories with low IoU scores. For example, our method obtains 10.4\% and 9.4\% relative improvements on \emph{Wall} than baseline and CWD, respectively. We further show the qualitative segmentation results visually in Fig.~\ref{Qualitative}. We can observe that our CIRKD produces more consistent semantic labels with the ground truth than baseline and CWD, indicating more meaningful pixel dependencies are captured.

T-SNE visualization of learned feature embeddings on the student network by CWD and our proposed CIRKD is shown in Fig.~\ref{tsne}. Compared to the CWD, the network trained by CIRKD shows a well-structured pixel-wise semantic feature space. The visual result suggests that learning cross-image pixel relations from the teacher network would help the student achieve better intra-class compactness and inter-class separability, thus improving segmentation performance.
\begin{table}[tbp]
	\centering
	\resizebox{1.\linewidth}{!}{
		\begin{tabular}{l|c|c|c}  
			\hline
			Method&Params (M)&FLOPs (G)&Test mIoU (\%) \\ 
			\hline
			T: DeepLabV3-Res101 &61.1M &280.2G &69.84 \\
			\hline
			S: DeepLabV3-Res18 &\multirow{5}{*}{13.6M} &\multirow{5}{*}{61.0G} & 66.92 \\
			+SKD~\cite{liu2019structured} & & & 67.46\\
			+IFVD~\cite{wang2020intra} & & &67.28 \\
			+CWD~\cite{shu2021channel} & & & 67.71 \\
			+CIRKD (ours) & & & \textbf{68.21} \\
			\hline
			S: PSPNet-Res18 &\multirow{5}{*}{12.9M} &\multirow{5}{*}{45.6G} &66.73 \\
			+SKD~\cite{liu2019structured} & & &67.83 \\
			+IFVD~\cite{wang2020intra} & & & 67.61\\
			+CWD~\cite{shu2021channel} & & & 67.92\\
			+CIRKD (ours) & & &\textbf{68.65} \\
			\hline
	\end{tabular}}
\vspace{-0.2cm}
	\caption{Performance comparison with state-of-the-art distillation methods over various student segmentation networks on CamVid. FLOPs is measured based on the test size of $360\times 480$. }
	\label{camvid} 
	\vspace{-0.1cm}
\end{table}

\subsubsection{Results on CamVid}
In Table~\ref{camvid}, we evaluate various distillation methods on CamVid. Our CIRKD achieves the best performance consistently. It outperforms the state-of-the-art CWD by 0.50\% and 0.73\% mIoU gains over DeepLabV3 and PSPNet, respectively.

\subsubsection{Results on Pascal VOC}
Beyond scene-parsing datasets, we also evaluate our CIRKD on Pascal VOC, a representative visual object segmentation dataset. As shown in Table~\ref{pascal_voc}, CIRKD achieves the best performance compared to other segmentation KD approaches. It surpasses the best completing CWD by 0.48\% and 0.79\% mIoU improvements on DeepLabV3 and PSPNet, respectively. The results demonstrate the scalability of our CIRKD to work reasonably well on visual object segmentation.

\begin{table}[tbp]
	\centering
	\resizebox{1.\linewidth}{!}{
		\begin{tabular}{l|c|c|c}  
			\hline
			Method&Params (M)&FLOPs (G)&Val mIoU (\%) \\ 
			\hline
			T: DeepLabV3-Res101 &61.1M &1294.6G &77.67 \\
			\hline
			S: DeepLabV3-Res18 &\multirow{5}{*}{13.6M} &\multirow{5}{*}{305.0G} & 73.21 \\
			+SKD~\cite{liu2019structured} & & & 73.51\\
			+IFVD~\cite{wang2020intra} & & &73.85 \\
			+CWD~\cite{shu2021channel} & & & 74.02 \\
			+CIRKD (ours) & & & \textbf{74.50} \\
			\hline
			S: PSPNet-Res18 &\multirow{5}{*}{12.9M} &\multirow{5}{*}{260.0G} &73.33 \\
			+SKD~\cite{liu2019structured} & & &74.07 \\
			+IFVD~\cite{wang2020intra} & & & 73.54\\
			+CWD~\cite{shu2021channel} & & & 73.99\\
			+CIRKD (ours) & & &\textbf{74.78} \\
			\hline
	\end{tabular}}
\vspace{-0.2cm}
	\caption{Performance comparison with state-of-the-art distillation methods over various student segmentation networks on Pascal VOC. We report the FLOPs based on the crop size of $512\times 512$ since the validation set does not have a fixed input size. }
	\label{pascal_voc} 
	
\end{table}

\subsection{Ablation Study and Parameter Analysis}
We conduct thorough ablation experiments of our proposed CIRKD on the Cityscapes validation set, a standard benchmark for semantic segmentation. For all experiments, we choose DeepLabV3-ResNet101 as the teacher and DeepLabV3-MobileNetV2 as the student by default.

\begin{table}[t]
	\centering
	\resizebox{1\linewidth}{!}{
		\begin{tabular}{l|c|cccccc}  
			\hline 
			Loss&Baseline& \multicolumn{6}{c}{Distillation}\\  
			\hline
			$L_{kd}$&-&\checkmark&\checkmark&\checkmark&\checkmark&\checkmark&\checkmark \\
			$L_{batch\_p2p}$&-&-&\checkmark&-&-&-&\checkmark\\
			$L_{memory\_p2p}$&-&-&-&\checkmark &-&\checkmark&\checkmark\\
			$L_{memory\_p2r}$&-&-&-&-&\checkmark&\checkmark&\checkmark\\
			\hline
			mIoU (\%) &73.12&74.26&74.87&75.11&74.94&75.26&\textbf{75.42}\\
			\hline
	\end{tabular}}
	\vspace{-0.2cm}
	\caption{Ablation study of distillation loss terms on Cityscapes \texttt{val}. Baseline denotes the cross-entropy loss $L_{task}$ (Equ.~(\ref{task})).}
	\label{ablation}
	\vspace{-0.2cm}
\end{table}

\textbf{Ablation study of loss terms.} As shown in Table~\ref{ablation}, we examine the contribution of each distillation loss. The conventional KD loss $L_{kd}$ improves the baseline by 1.14\%. Applying cross-image relational KD losses of $L_{batch\_p2p}$, $L_{memory\_p2p}$ and $L_{memory\_p2r}$ lead to 0.61\%, 0.85\% and 0.68\% mIoU gains over $L_{kd}$, respectively. The results show two conclusions: (1) Pixel-to-pixel distillation is more informative than the pixel-to-region counterpart. (2) Memory-based pixel-to-pixel distillation is better than the mini-batch-based counterpart, since the former can capture broader pixel dependencies from much more images than the latter. Finally, applying all losses together maximizes the segmentation performance, reducing the gap between the student and teacher from 4.95\% to 2.65\%.

\begin{figure}
	\centering 
	
	\begin{subfigure}[t]{0.238\textwidth}
		\centering
		\includegraphics[width=\textwidth]{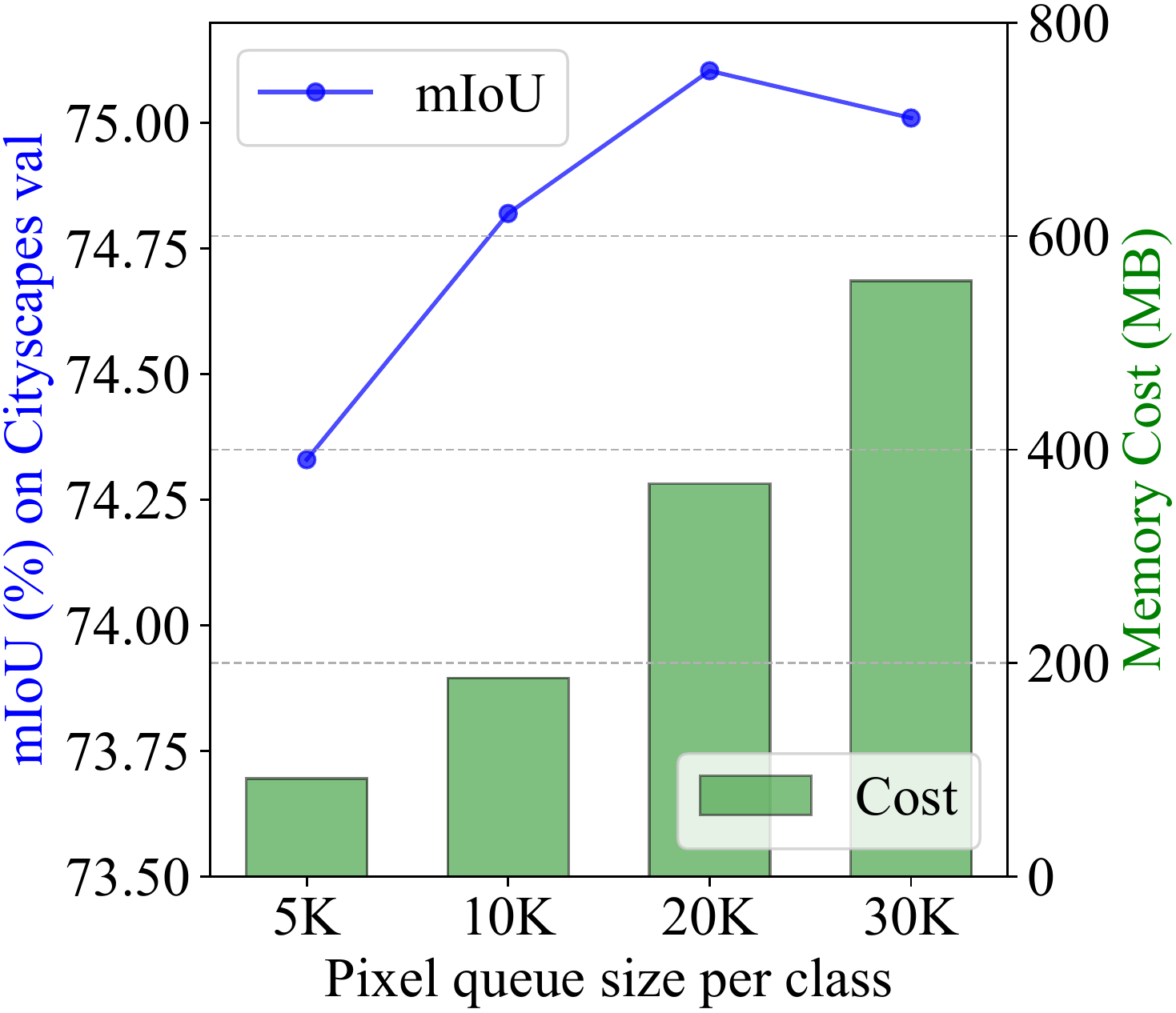}
		\caption{Pixel queue size $N_{p}$ per class}
		\label{pixel_queue}
	\end{subfigure}
	\begin{subfigure}[t]{0.234\textwidth}
		\centering
		\includegraphics[width=\textwidth]{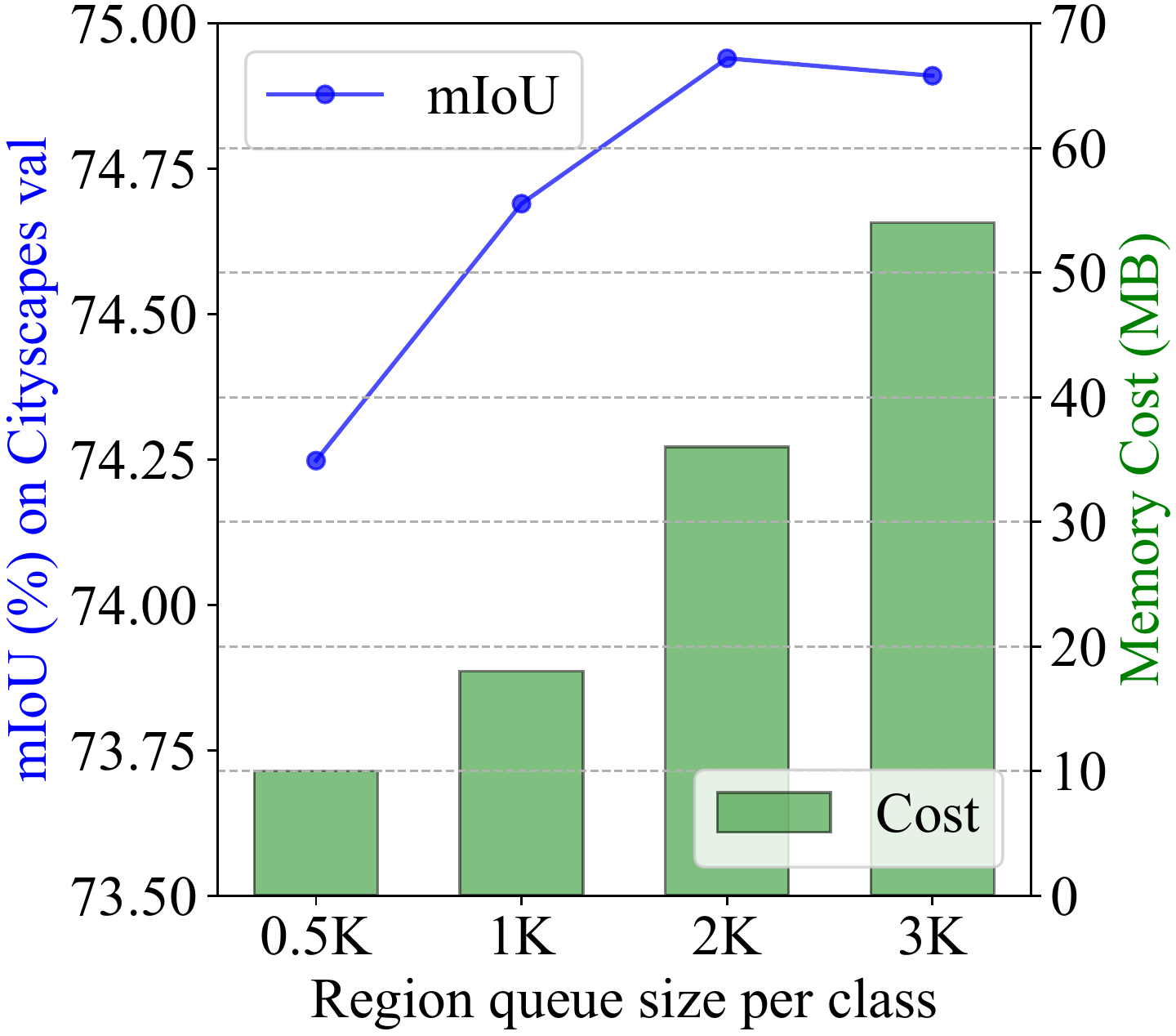}
		\caption{Region queue size $N_{r}$ per class}
		\label{region_queue}
	\end{subfigure}
	\vspace{-0.1cm}
	\caption{Impact of the (a) pixel queue size $N_{p}$ per class and (b) region queue size $N_{r}$ per class on Cityscapes \texttt{val}. 'Memory Cost' denotes the occupied GPU memory size.} 
	\label{memory} 
	
\end{figure}

\begin{figure}
	\centering 
	
	\begin{subfigure}[t]{0.18\textwidth}
		\centering
		\includegraphics[width=\textwidth]{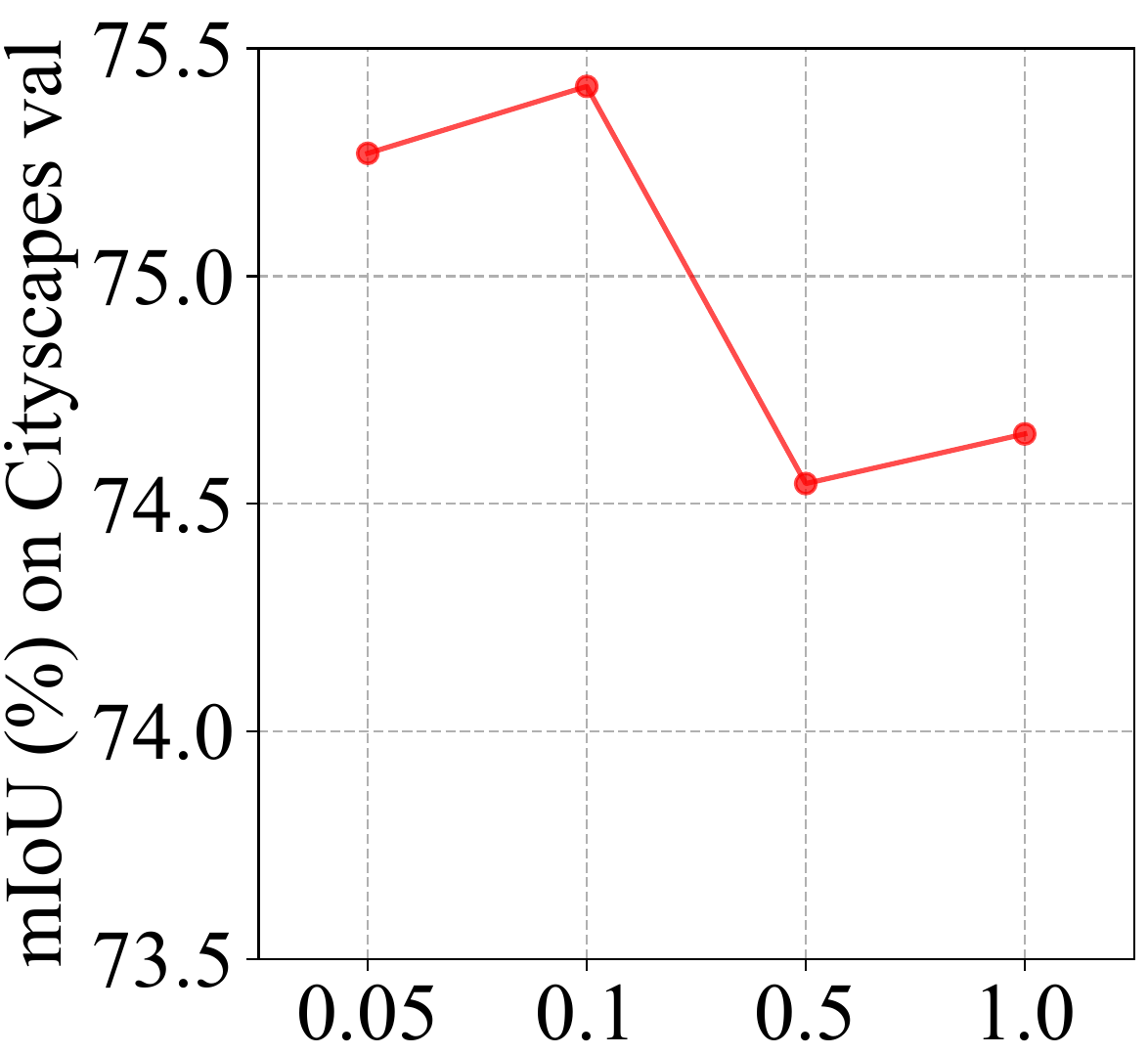}
		\caption{Temperature $\tau$}
		\label{temperature}
	\end{subfigure}
	\begin{subfigure}[t]{0.142\textwidth}
		\centering
		\includegraphics[width=\textwidth]{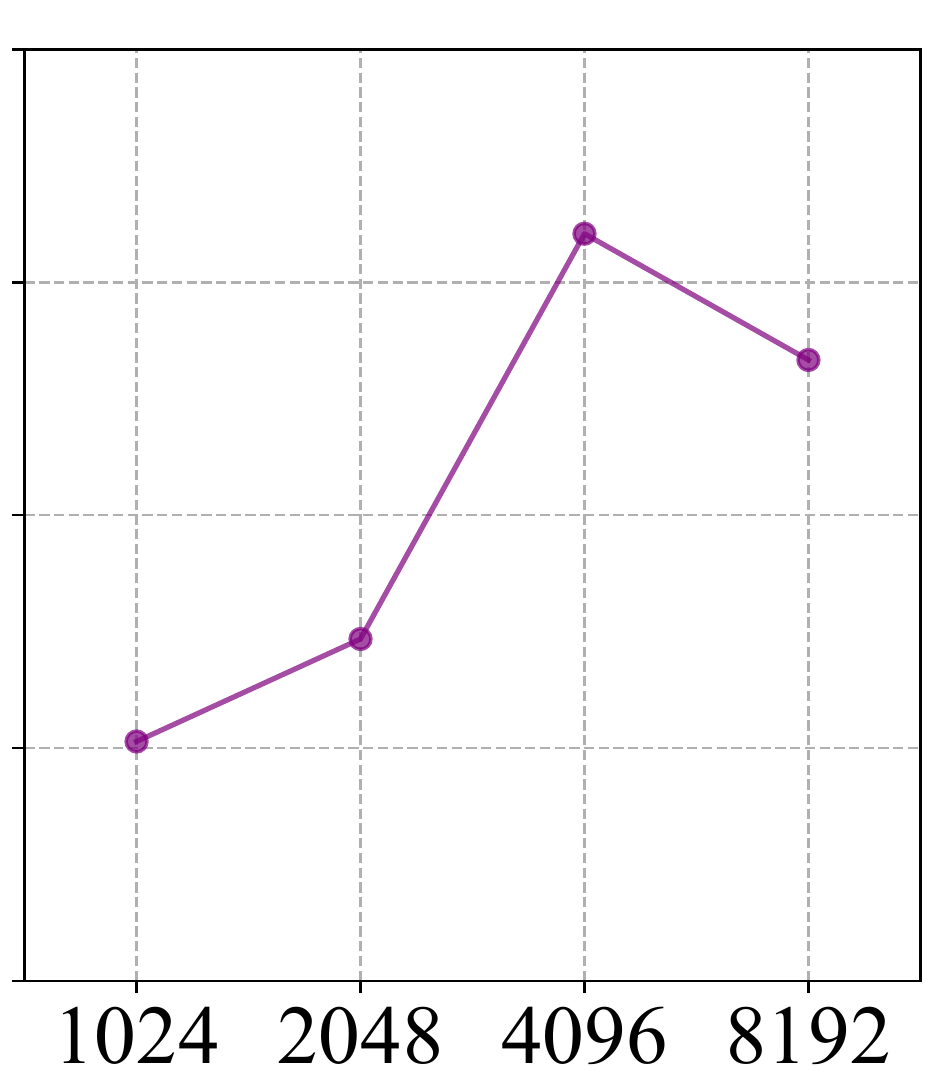}
		\caption{Pixel-based $K_{p}$}
		\label{Pixel}
	\end{subfigure}
	\begin{subfigure}[t]{0.142\textwidth}
		\centering
		\includegraphics[width=\textwidth]{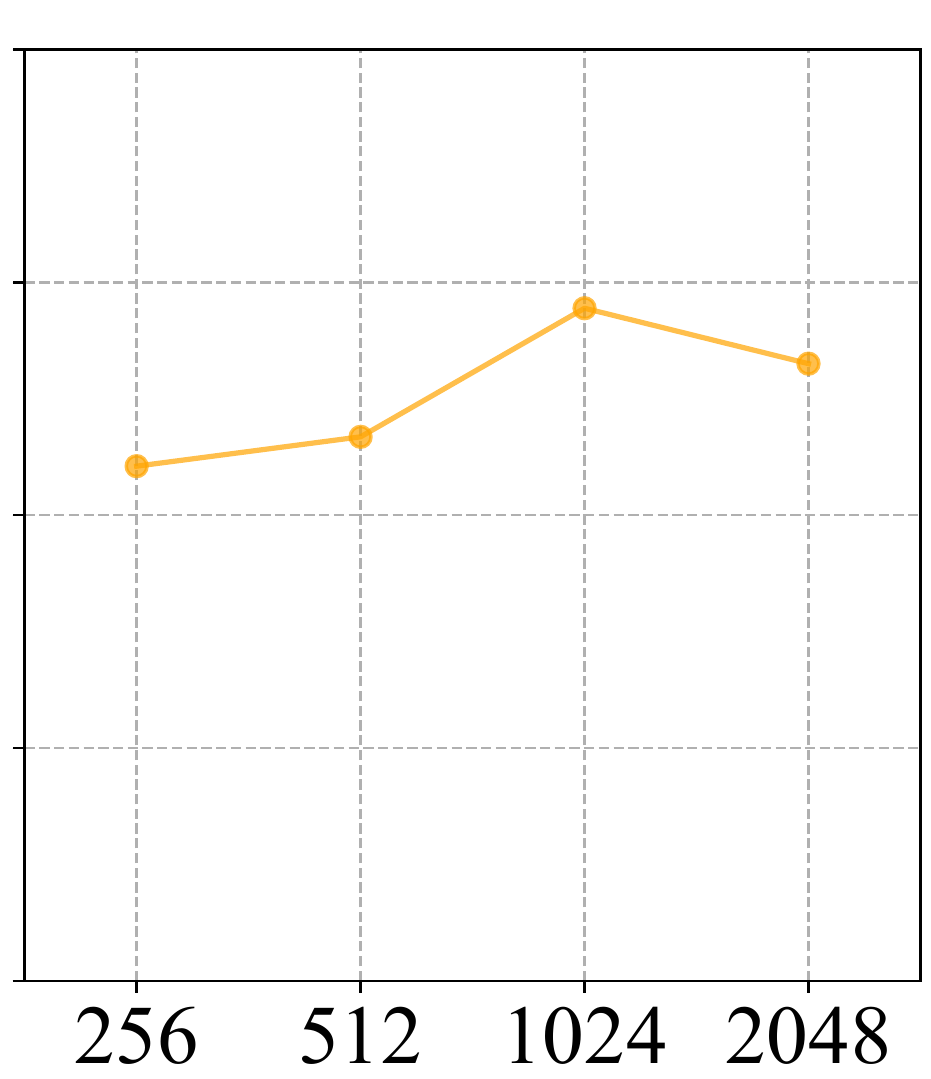}
		\caption{Region-based $K_{r}$}
		\label{Region}
	\end{subfigure}
	\label{params} 
	\vspace{-0.1cm}
	\caption{Impact of (a) the temperature $\tau$ and (b) the number of contrastive pixel embeddings $K_{p}$ and (c) the number of contrastive region embeddings $K_{r}$ on Cityscapes \texttt{val}.} 
	\vspace{-0.3cm}
\end{figure}

\textbf{Impact of the queue size.} We investigate the impact of memory sizes of the pixel queue and region queue. As shown in Fig.~\ref{memory}, distillation performance increases as the sizes of the pixel queue and region queue grow. This is because a larger queue could provide more abundant and diverse embeddings for capturing long-range dependencies. The results also show the distillation performance may also saturate at a certain memory capacity.

\textbf{Impact of the temperature $\tau$.} Temperature $\tau$ is used to calibrate the similarity distribution for relational KD. A more significant temperature $\tau$ brings a smoother distribution. As shown in Fig.~\ref{temperature}, we investigate the impact of $\tau$ in our CIRKD and find $\tau=0.1$ is the best choice.

\textbf{Impact of the number of contrastive embeddings.} As shown in Fig.~\ref{Pixel} and Fig.~\ref{Region}, we examine the number of contrastive embeddings to calculate pixel-to-pixel and pixel-to-region similarity matrices. The distillation performance increases as $K_{p}$ and $K_{r}$ grow, because the similarity distribution with a larger dimension would encode broader pixel dependencies. The upper bound of distillation performance may saturate at $K_{p}=4096$ for pixel-to-pixel distillation and $K_{r}=1024$ for pixel-to-region distillation.

\section{Conclusion}
This paper presents a novel cross-image relational KD to transfer global pixel correlations from the teacher to the student for semantic segmentation. Compared to previous KD approaches, our method helps students learn broader pixel dependencies from the teacher. Experiments on public segmentation datasets demonstrate the effectiveness of our CIRKD. We hope our work can inspire future research to explore global pixel relationships for segmentation KD.

\section{Appendix}
\begin{figure}[!h]  
	\centering 
	\includegraphics[width=1\linewidth]{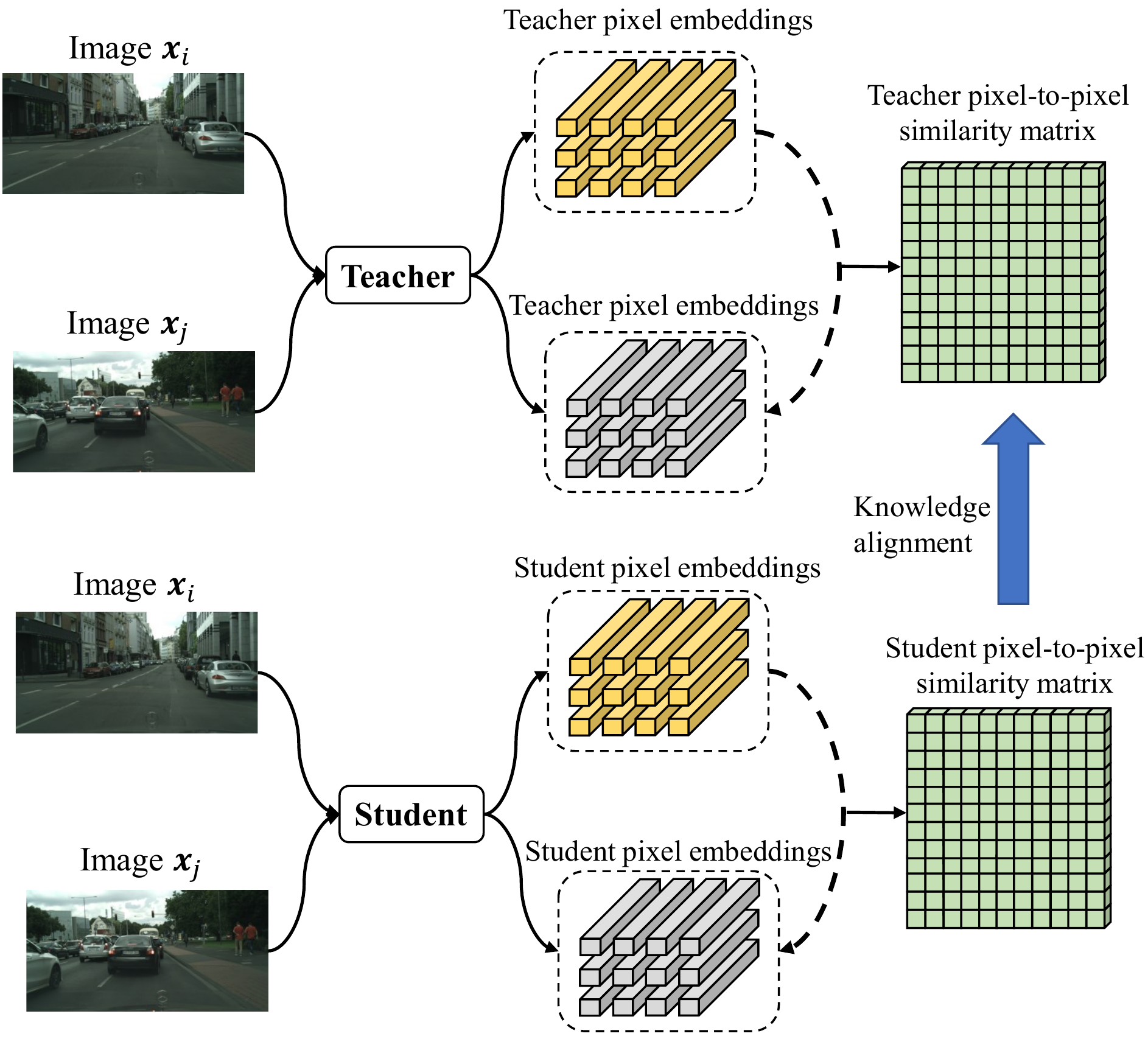}
	\caption{Overview of mini-batch based pixel-to-pixel distillation.} 
	\label{introduction}
\end{figure}
{\small
\bibliographystyle{ieee_fullname}
\bibliography{seg_egbib}
}

\end{document}